%% file: main.tex
\documentclass[letterpaper, 10 pt, journal, twoside]{IEEEtran}

\IEEEoverridecommandlockouts                              

\usepackage{graphics} 
\usepackage{epsfig} 
\usepackage{amsmath} 
\usepackage{amssymb}  
\usepackage{amsfonts}
\usepackage{optidef}
\usepackage{todonotes}
\usepackage{cite}
\usepackage{multicol}
\usepackage{multirow}
\usepackage{amsthm}
\usepackage{comment}
\usepackage[bookmarks=true]{hyperref}
\input{preamble}

\title{\LARGE \bf
Generalizable Spacecraft Trajectory Generation \\ via Multimodal Learning with Transformers
}

\author{Davide Celestini$^{1}$, Amirhossein Afsharrad$^{2,5}$, Daniele Gammelli$^{3}$, Tommaso Guffanti$^{3}$, Gioele Zardini$^{4}$, \\Sanjay Lall$^{2}$, Elisa Capello$^{1}$, Simone D'Amico$^{3}$, and Marco Pavone$^{3}$
\thanks{$^{1}$Department of Mechanical and Aerospace Engineering, Politecnico di Torino, 10129 Torino, Italy,
        {\tt\footnotesize \{davide.celestini, elisa.capello\}@polito.it}}%
\thanks{$^{2}$Department of Electrical Engineering, Stanford University, 94305, USA,
    {\tt\footnotesize\{afsharrad,lall\}@stanford.edu}}%
\thanks{$^{3}$Department of Aeronautics and Astronautics, Stanford University, 94305, USA,
    {\tt\footnotesize\{gammelli,tommaso,damicos, pavone\} @stanford.edu}}%
\thanks{$^{4}$Laboratory for Information and Decision Systems, Massachusetts Institute of Technology, 02139, USA,
    {\tt\footnotesize gzardini@mit.edu}}%
\thanks{$^{5}$Aktus AI, 94403, USA,
    {\tt\footnotesize amir@aktus.ai}}%
}

\begin{document}

\IEEEpubid{This work has been submitted to the IEEE for possible publication. Copyright may be transferred without notice, after which this version may no longer be accessible.}
\maketitle

\begin{abstract}


Effective trajectory generation is essential for reliable on-board spacecraft autonomy.
Among other approaches, learning-based warm-starting represents an appealing paradigm for solving the trajectory generation problem, effectively combining the benefits of optimization- and data-driven methods.
Current approaches for learning-based trajectory generation often focus on fixed, \textit{single-scenario} environments, where key scene characteristics, such as obstacle positions or final-time requirements, remain constant across problem instances.
However, practical trajectory generation requires the scenario to be frequently reconfigured, making the single-scenario approach a potentially impractical solution.
To address this challenge, we present a novel trajectory generation framework that generalizes across diverse problem configurations, by leveraging high-capacity transformer neural networks capable of learning from multimodal data sources.
Specifically, our approach integrates transformer-based neural network models into the trajectory optimization process, encoding both scene-level information (e.g., obstacle locations, initial and goal states) and trajectory-level constraints (e.g., time bounds, fuel consumption targets) via multimodal representations.
The transformer network then generates near-optimal initial guesses for non-convex optimization problems, significantly enhancing convergence speed and performance.
The framework is validated through extensive simulations and real-world experiments on a free-flyer platform, achieving up to 30\% cost improvement and 80 \% reduction in infeasible cases with respect to traditional approaches, and demonstrating robust generalization across diverse scenario variations. 

Project website and code: \href{https://acc25art.github.io}{https://acc25art.github.io}

\end{abstract}

\section{Introduction}
    
Future in-orbit servicing, assembly, manufacturing, and logistics operations will demand increasingly autonomous rendezvous capabilities.
Achieving reliable autonomy requires efficient trajectory optimization, enabling systems to compute state and control trajectories that simultaneously satisfy constraints and optimize mission objectives.
In the context of space operations, trajectory optimization must balance two conflicting requirements~\cite{NesnasEtAl2021,MalyutaEtAl2021}.
On one hand, autonomous rendezvous necessitates strict performance and safety guarantees.
On the other hand, computational resources are severely limited, as space-grade processors are orders of magnitude less powerful than their commercial, Earth-bound counterparts.
Such computational constraints are further are further exacerbated by the inherently non-convex nature of most relevant trajectory generation problems, making them hard to solve efficiently and reliably onboard an \gls{acr:av}~\cite{RuggieroEtAl2023, RuggieroEtAl2024}.

Given the aforementioned challenges, there is a growing interest in applying \gls{acr:ml} techniques to enhance traditional trajectory generation methods.
This interest in mainly motivated by two factors.
\IEEEpubidadjcol
First, learning-based techniques offer an appealing paradigm for solving problems characterized by stochastic spacecraft dynamics, multi-step decision-making, and potentially highly non-convex objectives \cite{silver2016mastering, ParkEtAl2024}. 
Second, the computational overhead of using trained \gls{acr:ml} models during inference is low, potentially compatible with the limited computational capabilities available onboard spacecrafts \cite{ieee_2019_computing, ParkEtAl2024}.
However, learning-based methods are often sensitive to distribution shifts in unpredictable ways, whereas optimization-based approaches are more readily characterized in terms of robustness and out-of-distribution behavior.
As a result, the deployment of learning-based methods in real-world, safety-critical applications--such as in-orbit \gls{acr:rpo}--has been limited.

Recent work has demonstrated that learning-based warm-starting of numerical optimization is an appealing approach to solving the trajectory generation problem.
In this paradigm, \gls{acr:ml} models provide a close-to-optimal initial guess for a downstream optimization problem, enabling hard constraint satisfaction while efficiently converging to a local optimum near the provided warm-start~\cite{GuffantiGammelliEtAl2024,CelestiniGammelliEtAl2024, art_ieeeaero25,BanerjeeEtAl2020}.
Nonetheless, current approaches are often limited to a \emph{single-scenario} setting, where most scenario characteristics and problem requirements, such as constraints on the spacecraft state (e.g., target waypoints, obstacles' and keep-out zones' configuration) or performance specifications (e.g., final time requirements), remain fixed across problem instatiations.
Consequently, the single-scenario approach limits the range of applications that can be addressed through learning-based trajectory optimization methods.

To overcome these limitations, this paper extends the framework introduced in~\cite{GuffantiGammelliEtAl2024, CelestiniGammelliEtAl2024} by endowing the transformer model to process comprehensive, multimodal scene representations.
By incorporating detailed information about spatial features, dynamic constraints, and performance requirements, the augmented model can generalize to diverse scenario variations.
The contributions of this paper are threefold:
\begin{itemize}
    \item We extend the \gls{acr:art} to handle scenario variations in both the physical environment and performance requirements.
    \item We investigate architectural components and design decisions within our framework, including the choice of representations for physical obstacles, dataset characteristics, and training schemes.
    \item We empirically validate our method through simulations and real-world experiments on a free-flyer platform. 
    Our results demonstrate that the proposed framework substantially improves the generalization capabilities of learning-based warm-starting methods, while enhancing the performance of standard trajectory optimization algorithms in terms of cost, runtime, and convergence rates.
\end{itemize}



\section{Related work}
Our work builds upon previous approaches that employ \gls{acr:AI} for spacecraft on-board autonomy \cite{IzzoEtAl2019, BanerjeeEtAl2020, HovellEtAl2021, FedericiEtAl2022, GuffantiGammelliEtAl2024, CelestiniGammelliEtAl2024}, and methods that exploit high-capacity neural network models for control~\cite{ChenEtAl2021, JannerEtAl2021, ChiEtAl2023}.
Prior work on \gls{acr:AI} for spacecraft autonomy can be broadly classified into two categories.
The first category focuses on learning representations for action policies, value functions, or reward models~\cite{IzzoEtAl2019, HovellEtAl2021, FedericiEtAl2024} using techniques from \gls{acr:rl} or \gls{acr:sl}. 
Approaches in this category often lack guarantees on performance and constraint satisfaction, and are hindered by the computational expense of simulating high-fidelity spacecraft dynamics, particularly in online \gls{acr:rl} settings.
The second category leverages learning-based components to warm-start sequential optimization solvers~\cite{BanerjeeEtAl2020, GuffantiGammelliEtAl2024, CelestiniGammelliEtAl2024}.
This strategy enables hard constraint satisfaction while achieving faster convergence to a local optimum in the neighborhood of the provided warm-start.
However, despite the theoretical appeal of warm-starting optimization solvers, existing methods are typically designed for a \emph{single-scenario} setting.
In this context, most key scene characteristics are assumed to be fixed, and the \gls{acr:ml} model receives only partial information regarding the spatial features and performance requirements related to a specific task.
In practice, this drastically reduces the robustness of the overall framework even in the face of minor problem reconfigurations (e.g., executing the same trajectory with a different final time constraint).
To address these limitations, this work extends~\cite{GuffantiGammelliEtAl2024, CelestiniGammelliEtAl2024, art_ieeeaero25} by endowing the transformer model with the ability to process diverse input modalities and more accurately represent a given trajectory optimization problem.
By doing so, we improve the \emph{adaptability} and \emph{robustness} of the framework to variations in problem settings.

Our method is closely related to recent efforts that leverage high-capacity generative models for control.
For instance, prior studies have demonstrated how transformers~\cite{ChenEtAl2021, JannerEtAl2021} and diffusion models~\cite{ChiEtAl2023}, trained via \gls{acr:sl} on pre-collected trajectory data, can be effective for model-free feedback control~\cite{ChenEtAl2021} and discrete model-based planning~\cite{JannerEtAl2021}.
However, such methods have two primary drawbacks.
First, they often ignore non-trivial state-dependent constraints, which are prevalent in practical applications and pose significant challenges for purely learning-based methods.
Second, they do not fully exploit the information available to system designers and practitioners, such as approximate knowledge of the system dynamics.

In contrast, our method addresses both of these shortcomings by (i) integrating online trajectory optimization to enforce non-trivial constraint satisfaction, and (ii) adopting a model-based approach in transformer-based trajectory generation leveraging available approximations of the system dynamics to improve autoregressive generation.


\section{Methodology}
\label{sec:methodology}

Let us consider the time-discrete \gls{acr:ocp}:
{\fontsize{9pt}{11.5pt} \selectfont
\begin{mini!}|l|[2]
    {\bx(t_i),\bu(t_i)}
    {\calJ = \sum_{i=1}^{N}{\cost\left(\bx(t_i), \bu(t_i)\right)}\label{opt:cost}}{\label{optimizationProblem}}{}
    \addConstraint{\bx(t_{i+1}) = \boldf\left(\bx(t_i), \bu(t_i)\right) \hspace{5mm} \forall i \in \left[1,N\right]}{\label{constr:dynamics}}
    \addConstraint{\bx(t_i) \in \calX_{t_i}, \, \bu(t_i) \in \calU_{t_i} \hspace{7.3mm} \forall i \in \left[1,N\right],}{\label{constr:equality}}
\end{mini!}}where~$\bx(t_i) \in \real^{\statedim}$ and~$\bu(t_i) \in \real^{\controldim}$ are the~$\statedim$-dimensional state and $\controldim$-dimensional control vectors, respectively,
$\cost : \real^{\statedim + \controldim} \rightarrow \real$ defines the cost function, $\boldf:\real^{\statedim+\controldim} \rightarrow \real^{\statedim}$ represents the system dynamics, $\calX_{t_i}$ and $\calU_{t_i}$ are the state and control constraint sets associated to the specific scenario of interest of the optimization problem, and where $\horizon \in \integer$ defines the number of discrete time instants ~$t_i$ over the full \gls{acr:ocp} horizon~$T_f$.
Our approach, as introduced in~\cite{GuffantiGammelliEtAl2024, CelestiniGammelliEtAl2024}, generates the state and control sequences~$\bX = \tup{\bx_1, \dots, \bx_N},\bU = \tup{\bu_1, \dots, \bu_N}$ by combining two components:
\begin{align}
\tup{\hat \bX, \hat \bU} & \sim p_{\theta}(\bX, \bU \, | \, \initcond), \label{eq:cond_prob}\\
\tup{\bX, \bU} & = \opt \left(\bx(t_1), \hat \bX, \hat \bU \right),
\end{align}
where~$p_{\params}(\cdot)$ denotes the conditional probability distribution over trajectories given an initial condition~$\initcond$, learned by a transformer model with parameters~$\params$,~$\opt$ denotes the trajectory optimization problem, and~$\hat \bX$,~$\hat \bU$ denote the predicted state and control trajectories provided as initial guesses to the optimization problem.
As discussed in the remainder of this section, the initial condition $\initcond$ is used to inform trajectory generation in the form of, e.g., an initial state~$\bx(t_1)$, a goal state~$\bx(t_N)$, scene descriptions, or other performance-related metrics (e.g., time constraints).

In this section, we first delve into the details of the multimodal trajectory representation employed for transformer training.
We then discuss the transformer architecture and training scheme devised to improve its effectiveness in generating close-to-optimal trajectories.

\subsection{Multimodal Trajectory Representation}
\label{subsec:multimodal_traj_gen}
A key element to enable reliable generalization to variations in the problem settings is the representation of trajectory data as sequences suitable for modeling by a transformer network~\cite{GuffantiGammelliEtAl2024}.
Let us consider a pre-collected dataset of trajectories in the form
    $\traj_{\mathrm{raw}} = \tup{\bx_1, \bu_1, r_1, \dots, \bx_N, \bu_N, r_N},$
where~$\bx_i$ and~$\bu_i$ denote the available state estimate and control signal at time~$t_i$, respectively, and~$r_i = - \cost(\bx(t_i), \bu(t_i))$ is the instantaneous reward (the negative of the cost function).
We define the following multimodal trajectory representation for transformer training:
\begin{equation}
    \label{eq:trajectory:representation}
    \traj = \tup{\bx_1, \calP_1, \calS_1, \bu_1, \dots, \bx_N, \calP_N, \calS_N, \bu_N},
\end{equation}
where~$\bx_1$ is the initial state,
and~$\calS_i, \calP_i$ are sets of \textit{scene} and \textit{performance} descriptors providing quantitative information regarding the scenario of interest for Problem (\ref{optimizationProblem}).
Both~$\calS$ and~$\calP$ can be represented through diverse modalities.
For instance,~$\calS$ may include various forms of environmental information, such as vector-space representations~$\calX_{\vectorspace}$ (e.g., positions and sizes of obstacles, waypoints, or other relevant spatial features), visual data~$\calX_{\image}$ (e.g., camera images or depth maps), and sensor readings~$\calX_{\sens}$.
Similarly,~$\calP$ may include diverse metrics that define the desired performance of the trajectory.
Specifically, we define~$\calG_i$ as the \textit{goal} or \textit{target state}, i.e., the state we wish to reach by the end of the trajectory (which can be either constant during the entire trajectory or time-dependent). 
Another performance metric of interest is the \textit{time-to-go}~$\calT_i$, expressed as the time remaining until the end of the trajectory, i.e., $\calT_i = T_f - t(i)$.
We further refer to~$\calR_i$ and~$\calC_i$ as the \textit{reward-to-go} and \textit{constraint-violation-budget} evaluated at~$t_i$, respectively.
Formally, as in~\cite{GuffantiGammelliEtAl2024}, we define~$\calR_i$ and ~$\calC_i$ to express future optimality and feasibility of the trajectory as:
\begin{equation}
    \label{eq:rtg}
    {\fontsize{9pt}{11.5pt} \selectfont \calR(t_i) = \sum_{j = i}^{\horizon}  r_j, \quad \quad C(t_i) = \sum_{j = i}^{N}{{\textsf{C}}(t_j)},}
\end{equation}
where~${\textsf{C}}(t_j)$~is an indicator function that checks for constraint violations at time~$t_j$:
\begin{equation}
\label{eq:ctg}
    {\fontsize{9pt}{11.5pt} \selectfont {\textsf{C}}(t_j) = 
\begin{cases}
    1, & \text{if } \exists \tup{\bx(t_j), \bu(t_j}) \notin \calX_{t_j}\times \calU_{t_j}, \\
    0, & \text{otherwise}.
\end{cases}}
\end{equation}
This enriched trajectory representation allows the transformer model to capture the causal relationships between states, scene and performance descriptors, and control profiles, thus representing a core design principle when targeting effective generalization across these dimensions.
Crucially, the representation defined in \cref{eq:trajectory:representation} maintains all the benefits discussed in~\cite{CelestiniGammelliEtAl2024}, namely: 
(i) during training, all performance parameters can be readily computed from raw trajectory data using \cref{eq:rtg} and \cref{eq:ctg} for~$\calR_i$ and~$\calC_i$, respectively, and by setting~$\calG_i = \bx_{\horizon}$, $\calT_1 = T_f$, and (ii) during inference, according to \cref{eq:cond_prob}, it allows the user to condition the generation of predicted state and control trajectories~$\hat \bX$ and~$\hat \bU$ through user-defined initial conditions $\initcond = \tup{\bx_1, \calP_1, \calS_1}$. 
For instance, given an initial state $\bx_1$ and corresponding scene description $\calS_1$, the user may query the transformer by defining a specific choice of $\calP_1$, e.g., a goal state, final time, cost, or combination thereof.

To clarify the notation introduced above, consider the following example:

\begin{example}[\textit{The Autonomous Rendezvous Transformer}]

Consider the scenario introduced in~\cite{GuffantiGammelliEtAl2024}, where a service spacecraft must rendezvous and dock with a target spacecraft or space station.
Specifically,~\cite{GuffantiGammelliEtAl2024} defines a dataset of trajectories~$\tau$ as:
\begin{equation}
    \tau = \tup{\bx_1, \calR_1, \calC_1, \bu_1 \ldots, \bx_N,, \calR_N, \calC_N},
\end{equation}
where~$\bx_1$ is the state of the spacecraft expressed using e.g., a relative Cartesian state, and~$\calR$ and~$\calC$ are the reward-to-go and constraint-violation-budget as defined in \cref{eq:rtg} and \cref{eq:ctg}, respectively.

According to the multimodal trajectory defined in \cref{eq:trajectory:representation}, the representation used in~\cite{GuffantiGammelliEtAl2024} may be recovered by defining the following quantities:
\begin{itemize}
    \item $\bx = \{\delta \boldsymbol{r}, \delta \boldsymbol{v} \} \in {\mathbb{R}}^6$, where $\delta \boldsymbol{r} = \{\delta r_r, \delta r_t, \delta r_n \} \in {\mathbb{R}}^3$ is the relative position and $\delta \boldsymbol{v} = \{\delta v_r, \delta v_t, \delta v_n \} \in {\mathbb{R}}^3$ is the relative velocity;
    \item $\bu = \Delta \boldsymbol{v} \in {\mathbb{R}}^3$, where $\Delta \boldsymbol{v}$ is the delta-V applied by the servicer spacecraft;
    \item $\calP = \{\calR, \calC\}$;
    \item $\calG = \{\emptyset\}, \calS = \{\emptyset\}, T_f \text{ is fixed}$.
\end{itemize}

\end{example}

\subsection{Transformer architecture for multimodal representation}
In this section, we introduce our proposed transformer architecture and the training scheme designed to handle the multimodal trajectory representation outlined in \cref{subsec:multimodal_traj_gen}. 
Our approach is based on two primary architectural components:

\noindent \textbf{Multimodal Encoder.}
To effectively process multimodal input sequences, the transformer leverages a three-step architecture.
First, each input modality is mapped to a \emph{shared} embedding space using modality-specific encoders.
Depending on the input type, such as camera images~$\calX_{\image}$, vector-field representations $\calX_{\vectorspace}$, etc., the encoder is represented by a differentiable mapping, which may be pre-trained (e.g., ResNet \cite{HeEtAl2016} backbone for image data).
Second, the resulting embeddings are processed by a GPT model with a causal attention mask~\cite{JannerEtAl2021}.
Given a maximum context length~$K$, the transformer uses the last~$K$ embeddings to update its internal representations.
Finally, the processed embeddings are mapped to the corresponding output domain via modality-specific decoders, which typically employ linear transformations.
Such transformations project the high-dimensional embeddings onto lower-dimensional outputs, such as state, control, or performance metrics.

\noindent \textbf{Diverse final times and variable-length sequences.} 
Generating state and control trajectories described by varying final times requires the transformer to produce output sequences of arbitrary lengths.
Drawing inspiration from standard practices in \gls{acr:llm} architectures, we train the transformer model on sequences of arbitrary lengths by randomly selecting sub-sequences of length~$K$ from the training data.
Once trained, the transformer can generate arbitrary sequence lengths by utilizing the last $K$ elements of each modality to autoregressively predict the next sequence element until the $N$-th element is reached.
This approach directly addresses the limitations of previous works~\cite{GuffantiGammelliEtAl2024, CelestiniGammelliEtAl2024}, which were restricted to fixed sequence lengths where~$K = N$.

\subsection{Pre-training}
\label{subsec:pre_training}
Let us introduce the training scheme used to learn the parameters $\theta$ of the transformer model.
At a high level, three main processes are required to obtain a functional transformer model: dataset generation, model training, and test-time inference.

\noindent\textbf{Dataset generation.}
The first step in our methodology involves generating a dataset suitable for effective transformer training.
To achieve this, we generate~$N_D$ trajectories by repeatedly solving diverse instances of the optimization problem described in \cref{optimizationProblem}.
The raw trajectories, along with their corresponding scene and performance descriptors, are subsequently re-arranged following the representation presented in \cref{eq:trajectory:representation}.

An essential property for effective generalization to unseen scenarios is dataset diversity.
Specifically, diversity on both scene and performance descriptors is crucial for enabling the transformer to learn the effects of various input modalities.
In practice, we propose to achieve this diversity through two strategies: randomization---across both initial conditions and scene descriptions---and problem relaxations.
When looking at randomization, a simple and effective strategy is to uniformly sample initial conditions (e.g., initial states), scene descriptors (e.g., the position of the obstacles), and performance descriptors (e.g., final time specifications) within a set of operating conditions of interest.
Moreover, to achieve diversity in cost and constraint violation performance, our datasets include both solutions to Problem (\ref{optimizationProblem}) and its relaxations (e.g., by relaxing obstacle avoidance constraints).
For instance, solutions to the relaxed versions of Problem (\ref{optimizationProblem}) typically yield trajectories with lower cost but non-zero constraint violations (i.e., higher~$\calR$ and~$\calC>0$). 
In contrast, direct solutions to Problem (\ref{optimizationProblem}) are characterized by higher cost and zero constraint violations ( i.e., lower $\calR$ and $\calC = 0$).
By including a broad range of behaviors specified by different scene and performance specifications, the transformer can effectively learn how these parameters influence trajectory generation.

\noindent\textbf{Training.}
We train the transformer by employing the standard teacher-forcing procedure commonly used in training sequence models.
Specifically, we optimize the following loss function:
\begin{align}
    \label{eq:ol_loss_function}
    \olloss(\tau) = \sum_{n=1}^{N_D} \sum_{i=1}^{N} \quad & \olloss(\bx_i^{(n)}, \hat \bx_i^{(n)}) + \olloss(\bu_i^{(n)}, \hat \bu_i^{(n)}) + \\
    & \olloss(\calP_i^{(n)}, \hat \calP_i^{(n)}) + \olloss(\calS_i^{(n)}, \hat \calS_i^{(n)}), \nonumber
\end{align}
where~$n$ indexes the~$n$-th trajectory sample from the dataset, $\hat{\bx}^{(n)}_i \sim p_{\params}( \bx^{(n)}_i \mid \tau^{(n)}_{<t_i}), \hat{\bu}^{(n)}_i \sim p_{\params}(\bu^{(n)}_i \mid \tau^{(n)}_{<t_i}, \bx^{(n)}_i)$ are the one-step predictions for both state and control vectors, $\hat \calP$ and $\hat \calS$ are the predicted performance and scene descriptors, respectively, and where we use $\tau_{<t_i}$ to denote a trajectory spanning from time steps $t \in \left[t_1, t_{i-1}\right]$.
While \cref{eq:ol_loss_function} provides a general definition for an effective loss function, the implementation specifics are necessarily scenario-dependent.
For instance, if in some cases predicting performance or scene elements may result in informative learning tasks (e.g., predicting the optimal final time given an initial state, or the evolution of moving obstacles in the scene), in other circumstances, the loss components defined in \cref{eq:ol_loss_function} may not need to be simultaneously active.

\noindent\textbf{Inference.}
Once trained, the transformer can be used to autoregressively generate an open-loop trajectory, denoted as $(\hat{\bX},\hat{\bU}) \sim p_{\params}(\bX, \bU \, | \, \initcond)$, from a given initial condition  $\initcond = \tup{\bx_1, \calP_1, \calS_1}$.
Here,~$\initcond$ encodes the initial state together with scene and performance descriptors.
Given $\initcond$, the autoregressive generation process is defined as follows:
(i) the transformer generates a control input $\bu_1$ based on the initial conditions, 
(ii) inspired by the approach in~\cite{GuffantiGammelliEtAl2024}, we adopt a model-based autoregressive generation method.
The next state~$\bx_2$ is computed using a known approximate dynamics model $\dynapprox(\bx, \bu)$, which is a reasonable assumption for the problem setting we investigate, (iii) the performance and scene descriptors are updated, and (iv) the previous three steps are repeated for subsequent time steps until the desired time horizon is reached.
For instance, let us denote the performance descriptor as a tuple described by a goal state, reward-to-go, constraint-violation-budget, and time-to-go specification, i.e., $\calP = \tup{\calG, \calR, \calC, \calT}$. 
The update of the performance descriptors defined in step (iii) is achieved by decreasing the reward-to-go $\calR_1$, constraint-violation-budget $\calC_1$, and time-to-go $\calT_1$ by the instantaneous reward $r_1$, constraint violation ${\textsf{C}}(t_1)$, and time instant $\Delta t = T_f/N$, respectively, and defining the goal state $\calG_1$ to be either kept constant or updated to a new goal state.
In practice, we select $\calR_1$ as a (negative) quantifiable lower bound of the optimal cost and $\calC_1 = 0$, incentivizing the generation of near-optimal and feasible trajectories.

\section{Experiments}
\label{sec:experiments}
The goal of our experimental evaluation is to address the following key questions: (1) Can \gls{acr:art} generate effective warm-starts for trajectory optimization in novel scenarios? 
(2) How does data diversity influence \gls{acr:art}'s ability to generalize across different conditions?
(3) What are the critical design choices necessary to ensure robust generalization?

In this section, we evaluate the performance of our proposed approach using a \emph{free-flyer} testbed.
We first describe the free-flyer dynamics and problem specifications, followed by a detailed explanation of the experimental setup and the transformer architecture implemented.
We then present simulation results that demonstrate statistically significant improvements in trajectory planning across a range of time constraints and operational scenarios.
Finally, we provide an analysis of selected trajectories executed on the real robotic platform.

\subsection{Free-flyer dynamics and problem specification}
\label{subsec:freeflyer_problem}
\begin{figure}[tb]
\centering
\includegraphics[width=1\linewidth]{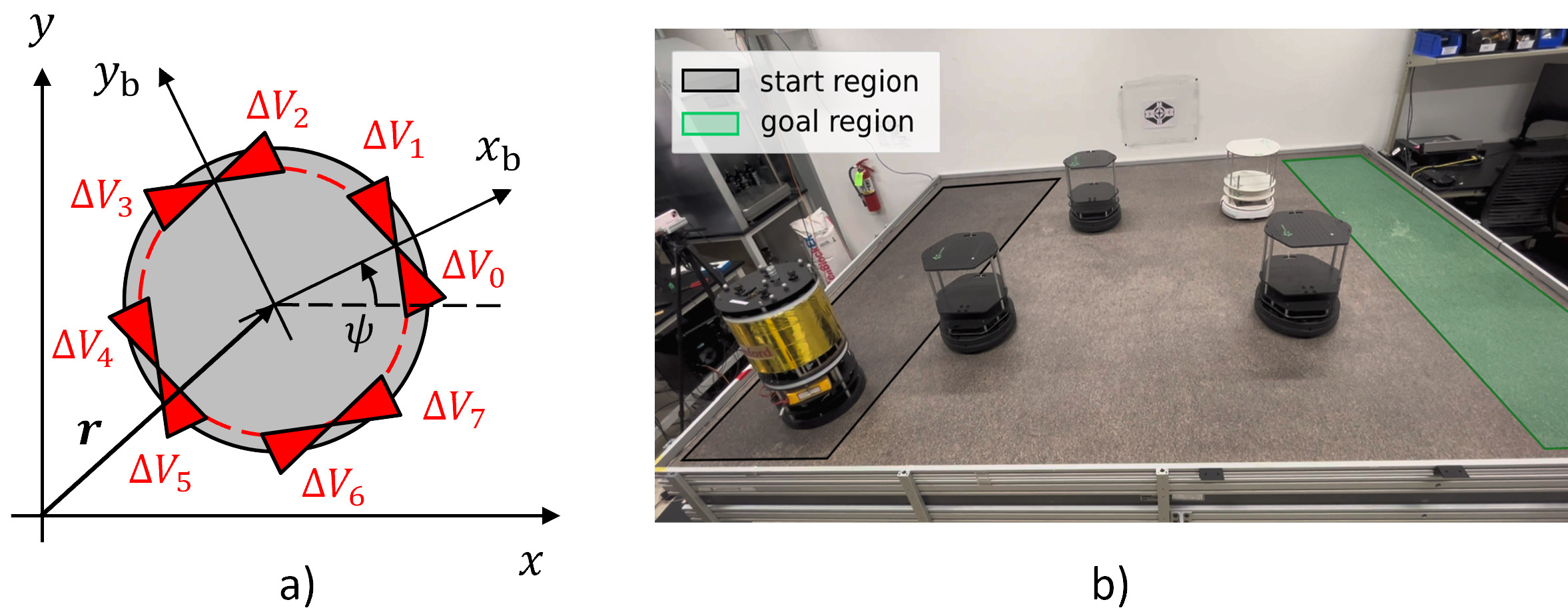}
    \caption{(a) Schematic representation of the free-flyer in the global reference frame~$\globalRF$. The eight-thruster configuration allows for independent control of rotational and translational motion. (b) A top-view of the free-flyer platform with highlighted start (black) and goal (green) regions.}
    \label{fig:freeflyer}
\end{figure}

We consider a free-flyer system designed to emulate orbital rendezvous maneuvers in a two-dimensional plane~\cite{CelestiniGammelliEtAl2024}.
The system is equipped with eight on-off thrusters arranged to provide independent control over both translational and rotational motion, as illustrated in \cref{fig:freeflyer}a).
By leveraging compressed air, the system achieves near-frictionless movement over a granite surface.

Let~$\globalRF$ denote the global Cartesian reference frame. 
The state vector of the system is defined as~$\bx:=\tup{\br,\psi,\bv,\omega} \in \real^6$, where~$\br,\bv \in \real^2$, $\psi, \omega \in \real$ are the position, velocity, heading angle and angular rate of the free-flyer, respectively.
Furthermore, the control input vector in the global reference frame is given by~$\bu:=\mat{R}_{\mathrm{GB}} (\psi) \Lambda \Delta \bV \in \real^3$, where~$\Delta \bV \in \real^8$ represents the impulsive delta-Vs applied by each thruster, $\Lambda \in \real^{3 \times 8}$ is the thruster configuration matrix, and $\mat{R}_{\mathrm{GB}} \in \real^{3 \times 3}$ is the rotation matrix from the body-fixed reference frame of the free-flyer $\bodyRF$ to the global frame~$\globalRF$.
Due to actuation limits, the delta-Vs applied by each thruster  are limited to $0 \le \Delta \bV \le \Delta \bV_{\mathrm{max}}$, where $\Delta \bV_{\mathrm{max}} = T \Delta t/m$, and $T$, $\Delta t$ and $m$ denote respectively the thrust level of the thrusters, the timestep used for time discretization and the mass of the free-flyer.
However, exploiting the aforementioned formulation of $\bu$, the dynamics of the system are modeled through the linear impulsive model $\bx_{i+1} = \bx_i + \mathrm{diag}\left([\Delta t, \Delta t, \Delta t, 1, 1, 1]\right) \bu_i$.

Our control objective is to design a sequence of control inputs~$\bu(t)$ that steer the system from an initial state~$\bx(t_1)=\bx_{\mathrm{start}}$ to a goal state~$\bx(T_f)=\bx_{\mathrm{goal}}$ within a specified time horizon~$[t_1,T_f]$, while satisfying the system dynamics and avoiding obstacles.
This problem is formulated as an \gls{acr:ocp}:
\begin{mini!}|s|[2]
    {\bx_i,\bu_i}
    {\sum_{i=1}^{N}{\lVert \bu_i \rVert_1}\label{opt2:cost}
    }{\label{optimizationProblem2}}{}
    \addConstraint{\bx_{i+1} = \dynapprox\left(\bx_i, \bu_i\right) \hspace{19.5mm} \forall i \in \left[1,N\right]}{\label{constr2:dynamics}}
    \addConstraint{\bx_1=\bx_{\start}}{\label{constr2:initial}}
    \addConstraint{\bx_{N+1}=\bx_{\goal}}{\label{constr2:final_state}}
    \addConstraint{\bx_i \in \calX_{\mathrm{table}} \hspace{29.5mm} \forall i \in \left[1,N\right]}{\label{constr2:state_limits}}
    \addConstraint{\Gamma \left( \bx_i \right) \ge 0 \hspace{30.5mm} \forall i \in \left[1,N\right]}{\label{constr2:koz}}
    \addConstraint{0 \le \Lambda^{-1}R_{\mathrm{GB}_i}^{-1}\bu_i \le \Delta \bV_{\mathrm{max}} \hspace{5.5mm} \forall i \in \left[1,N\right],}{\label{constr2:action_limits}}
\end{mini!}
where \cref{opt2:cost} expresses the minimization the control effort for impulsive thrusters, with~$N=T_f/\Delta t$ being the number of discrete time instants over the desired \gls{acr:ocp} horizon~$T_f$, \cref{constr2:dynamics} represents the system dynamics expressed in $\globalRF$, \cref{constr2:initial} and \cref{constr2:final_state} represent the initial state and the goal, and \cref{constr2:state_limits} defines the boundaries of the test bed. 
The non-convexity of the \gls{acr:ocp} is rooted in the obstacle avoidance constraint in \cref{constr2:koz}, caused by the nonlinear distance function~$\Gamma \colon \real^{m \times n_{x}}\! \rightarrow\! \real^{m}$ with respect to $m \in [1, M]$ non-convex keep-out-zones, and by the actuation limits in \cref{constr2:action_limits}, due to the nonlinear coupling between the free-flyer orientation and the shooting direction of its thrusts.

\subsection{Experimental design}
\label{subsec:experimental_design}
In our experiments, we aim to isolate (i) the benefits of the initial guess, and (ii) the effects of various design decisions on generalization.
Therefore, we keep the formulation and solution algorithm for the \gls{acr:ocp} fixed across scenarios. 
Specifically, we leverage \gls{acr:scp} to solve the trajectory optimization problem and evaluate the benefits of different warm-starting approaches.
For all simulation experiments, we compare our approach, \gls{acr:art}, with an \gls{acr:scp} approach (\gls{acr:scp} REL) where the initial guess to Problem (\ref{optimizationProblem2}) is provided by the solution to a relaxed version of the problem (REL) which represents a (potentially infeasible) cost lower bound to the full problem ~\cite{BanerjeeEtAl2020,AlcanKyrki2022}.
To assess the impact of specific algorithmic components, we perform several ablation studies focusing on dataset characteristics, choice of representation, and training scheme (\cref{tab:training_diversity}).
Throughout the experiments, trajectory generation performance is evaluated using three primary metrics: cost, \gls{acr:scp} convergence speed, and infeasibility rate.

\noindent\textbf{Transformer architecture.} In our experiments, we use a causal GPT model consisting of six layers, six attention heads, and an embedding dimension of $384$, as in~\cite{GuffantiGammelliEtAl2024, CelestiniGammelliEtAl2024}.
Depending on the experimental setup, we define the following loss functions for transformer training:
\begin{align}
    \label{eq:specialized_ol_loss_function}
    \olloss(\tau)^{\FT} = \sum_{n=1}^{N_D} \sum_{i=1}^{N} & \left( \lVert \bx^{(n)}_i - \hat{\bx}^{(n)}_i \rVert_2^2 + \lVert \bu^{(n)}_i - \hat{\bu}^{(n)}_i \rVert_2^2 + \right.\nonumber\\
    & \left. + \lVert \calT^{(n)}_i - \hat{\calT}^{(n)}_i \rVert_2^2 \right), \\
    \olloss(\tau)^{\OA} = \sum_{n=1}^{N_D} \sum_{i=1}^{N} & \left( \lVert \bx^{(n)}_i - \hat{\bx}^{(n)}_i \rVert_2^2 + \lVert \bu^{(n)}_i - \hat{\bu}^{(n)}_i \rVert_2^2 \right),
\end{align}
where $\lVert \cdot \rVert_2$ denotes the L2-norm, and $\FT$ and $\OA$ refer to the scenarios described by variations in $\mathrm{F}$inal $\mathrm{T}$ime or $\mathrm{O}$bstacle configurations.

\subsection{Simulation experiments}
\label{subsec:simulation_experiments}
\begin{table}[!t]
    \caption{ART models used to study the effect of dataset diversity.}
    \label{tab:training_diversity}
    \centering
    \resizebox{\columnwidth}{!}{
    \begin{tabular}{c|c|c|c}
        \hline
        \hline
        \multirow{2}{*}{Model} & Final times in & N. samples & N. scenarios \\
        & dataset $[s]$ & in dataset & in dataset \\
        \hline
        ART-1t & $70$ & $\approx 200,000$ & $1$\\
        ART-4t & $[40,60,80,100]$ & $\approx 200,000$ & $1$\\
        ART-Rt & $\mathrm{rand}(40,100)$ & $\approx 200,000$ & $1$\\
        \hline
        ART-1s & $40$ & $\approx 200,000$ & $1$\\
        ART-15s & $40$ & $\approx 800,000$ & $15$\\
        ART-Rs & $40$ & $\approx 800,000$ & $\mathrm{rand}(\calX_{   \mathrm{table}})$\\
        \hline
        \hline
    \end{tabular}}
\end{table}
In this section, through extensive simulation, we answer the three primary questions introduced in~\cref{sec:experiments}.

\vspace{1mm}
\noindent\textbf{(1) Can ART generate effective warm-starts for trajectory optimization in novel scenarios?}

To determine whether the proposed architecture can learn near-optimal behavior and generalize to novel scenarios, we first evaluate the performance of warm-starting the \gls{acr:scp} with \gls{acr:art} models trained on datasets whereby the scene descriptors (i.e., obstacle configurations and final time specifications) are uniformly sampled from a set of operating conditions of interest. 
We refer to these models as ART-Rs and ART-Rt for obstacle configurations and final time specifications, respectively. 
The comparisons are made by varying \emph{either} the final time \emph{or} the operational scenario.
In the former case, evaluations occur at seven different final times $T_f$, ranging from $40 \ s$ to $100 \ s$ in increments of $10 \ s$. 
In the latter case, the models are assessed using three testing scenarios, not seen during training. For each final time \textit{or} scenario, the testing dataset comprises $5,000$ problem specifications generated by randomly selecting $\bx_{\mathrm{start}}$ and $\bx_{\mathrm{goal}}$ from the predefined start and goal regions shown in \cref{fig:freeflyer}b), and by computing $N$ (i.e., the number of discrete time steps that characterize the trajectory) according to $T_f$ (i.e., the final time) and $\Delta t = 0.5 \ s$. 
To initialize the trajectory generation process, we set the performance parameter $\calR_1$ to the negative cost of the REL solution and $\calC_1 = 0$, with transformer context size $K = 100$.
\cref{tab:generalization_capabilities} summarizes the average percentage improvement in cost suboptimality and number of SCP iterations, as well as the reduction in infeasible cases compared to the baseline SCP REL. 
Both ART-Rs and ART-Rt exhibit significant generalization capabilities to unseen scenarios, outperforming SCP REL by substantially reducing the number of infeasible cases by over $70 \%$ and the suboptimality gap in the SCP solution by approximately $10 \%$.
\begin{table}[!t]
    \caption{ART-Rs and ART-Rt comparison with the baseline under generalized obstacle avoidance and time constraints.}
    \label{tab:generalization_capabilities}
    \centering
    \resizebox{\columnwidth}{!}{
    \begin{tabular}{c|c|c|c}
        \hline
        \hline
        \multirow{2}{*}{Warm-start} & Suboptimality & N. iterations & Infeasible cases\\
        & improvement & improvement & reduction \\
        \hline
        ART-Rs & $+8.74 \%$ & $+7.79 \%$ & $-81.99 \%$\\
        ART-Rt & $+13.56 \%$ & $+1.52 \%$ & $-78.58 \%$\\
        \hline
        \hline
    \end{tabular}}
\end{table}
\begin{figure}[tb]
\centering
\includegraphics[width=1\linewidth]{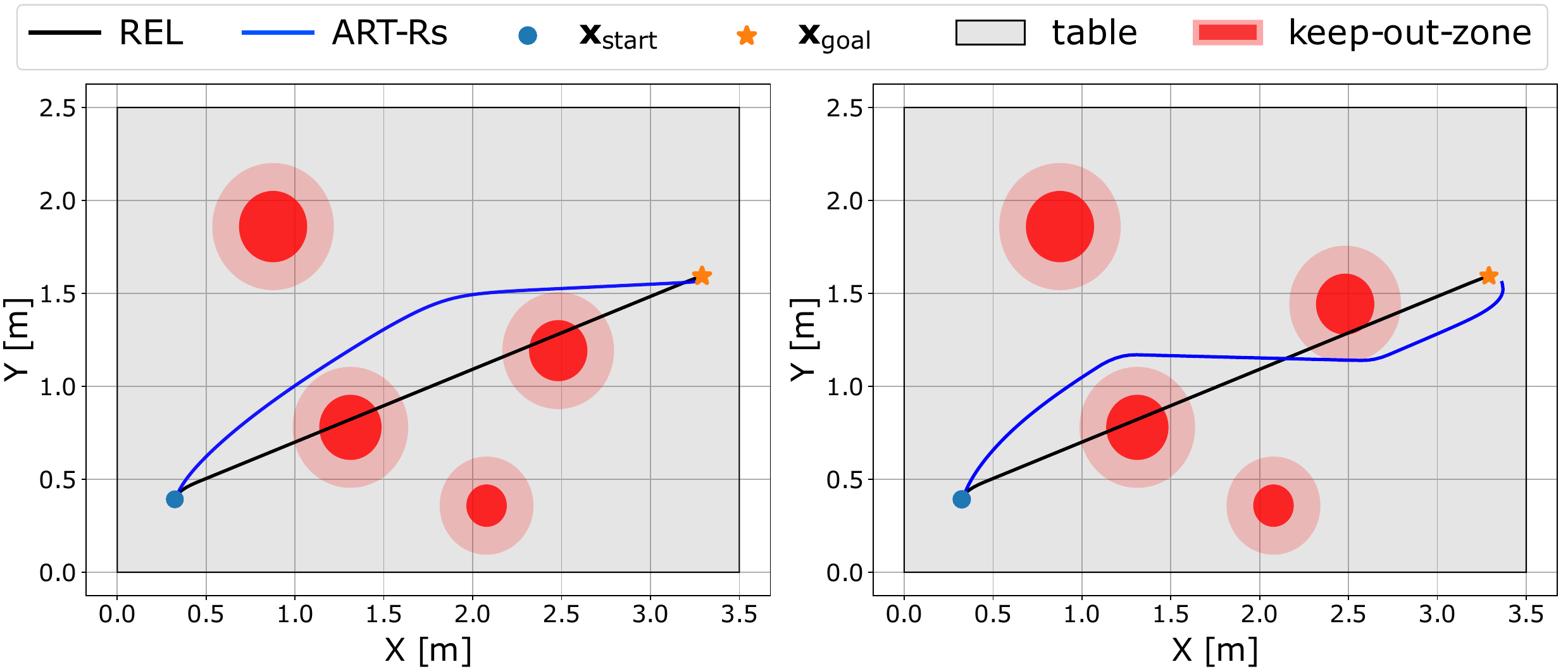}
    \caption{Comparison between warm-starting trajectories generated by ART-Rs and REL in scenarios with varying obstacle configurations. The modification of the right-most obstacle prompts ART to adapt its trajectory accordingly, demonstrating coherent behavior in response to the change.}
    \label{fig:obs_avoidance_ART}
\end{figure}
The generalization capabilities of the models can also be qualitatively assessed in \cref{fig:obs_avoidance_ART}, whereby we use ART-Rs to generate a warm-starting trajectory in two variations of the same scenario. 
As clearly noticeable in the figure, ART-Rs is able to correctly perceive the position of the right-most obstacle and accordingly plan a close-to-optimal warm-starting trajectory. 
On the contrary, the warm-start provided by REL is not affected by the obstacles' features.

\vspace{1mm}
\noindent \textbf{(2) How does data diversity influence ART's ability to generalize across different conditions?}

\begin{figure}[tb]
\centering
\includegraphics[width=1\linewidth]{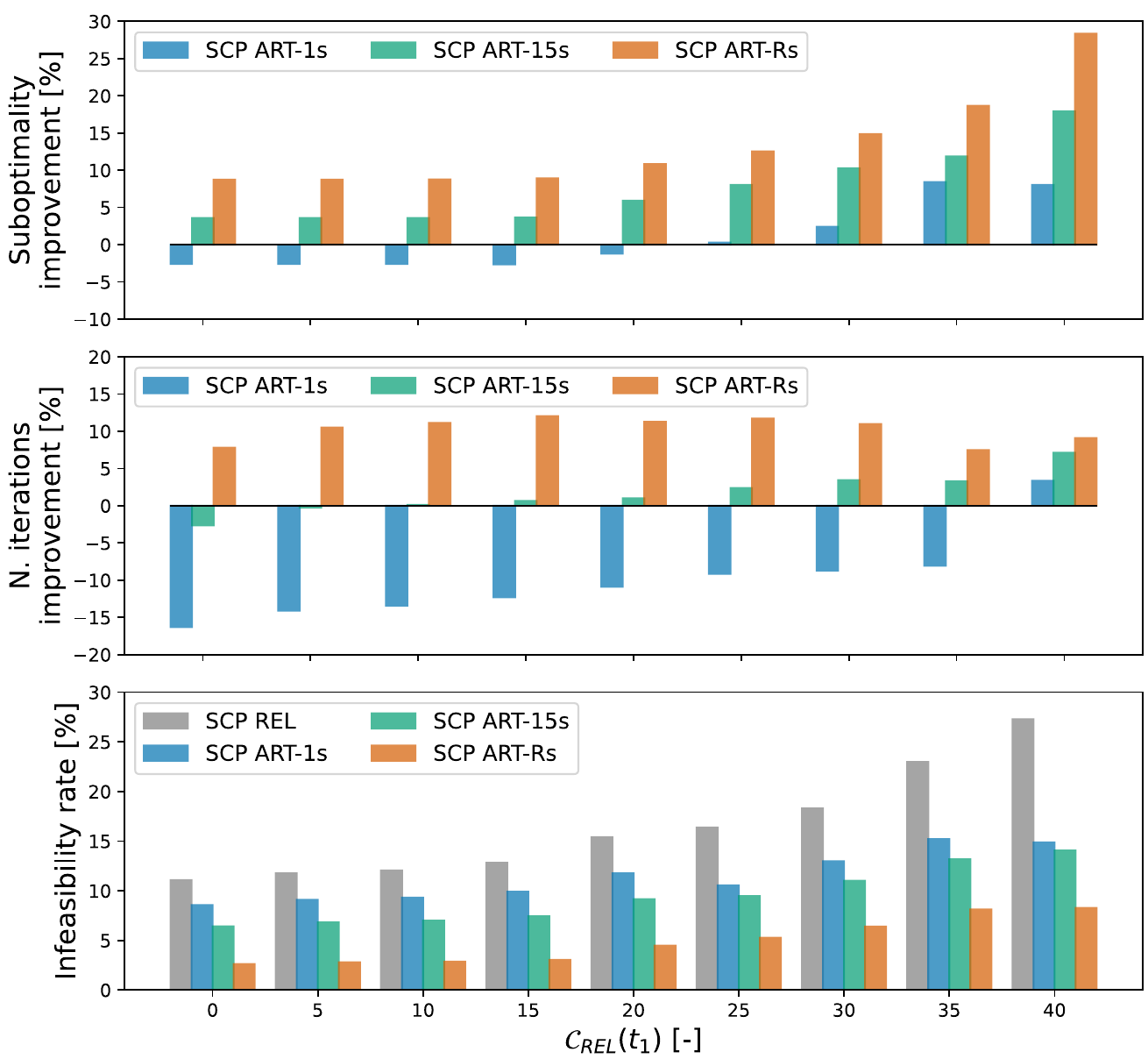}
    \caption{
    Percentage improvements in cost suboptimality (top) and number of SCP iterations (middle) relative to REL, along with a comparison of infeasibility rates (bottom), achieved by warm-starting the SCP using ART models trained on datasets with varying degrees of scenario diversity. 
    Each bar represents the average improvement for constraint-violation-budgets (i.e., $\mathcal{C}_\mathrm{REL} (t_1)$) greater than or equal to the corresponding x-axis value.}
    \label{fig:obstacles_diversity}
\end{figure}

\begin{figure}[tb]
\centering
\includegraphics[width=0.99\linewidth]{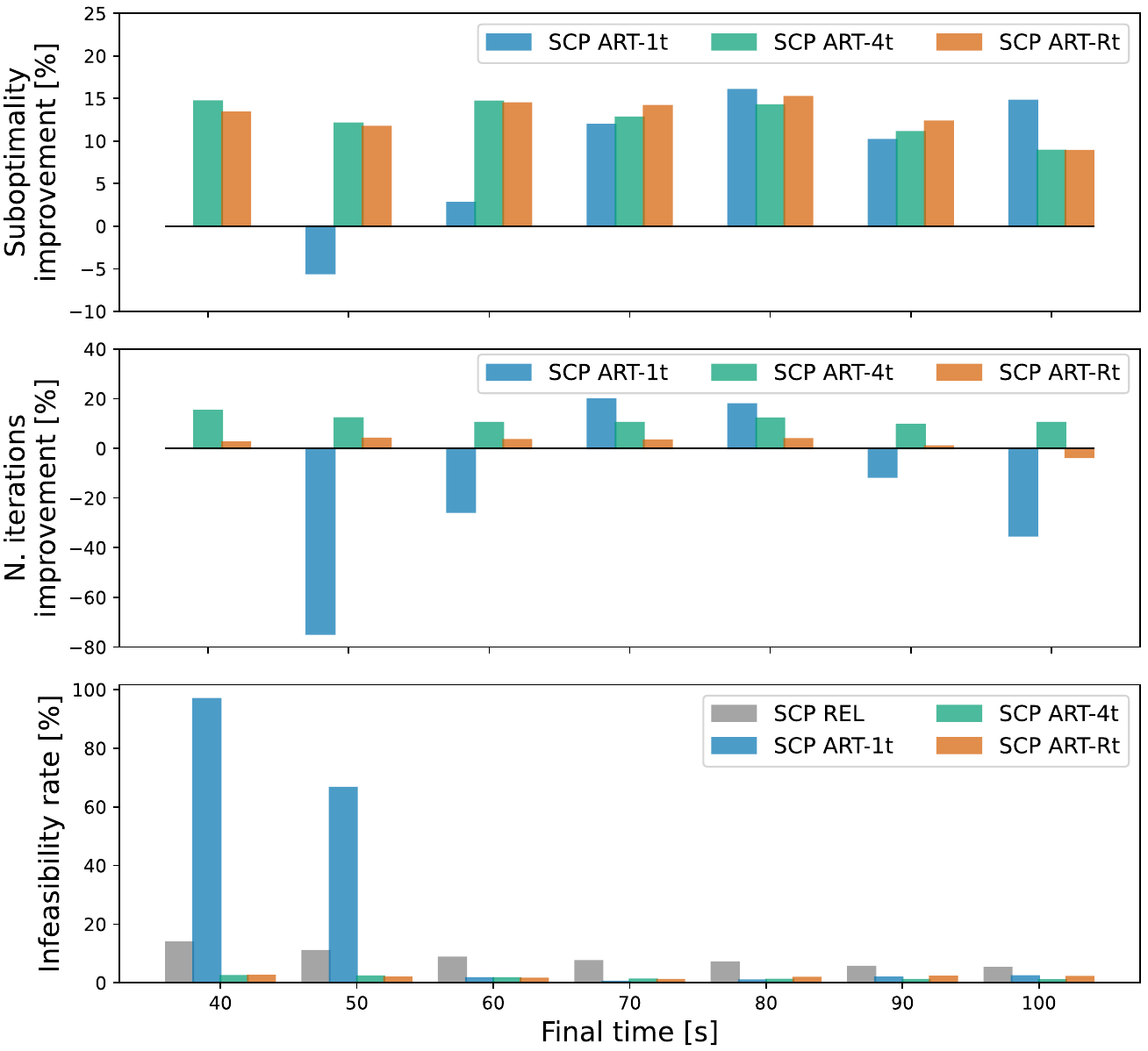}
    \caption{
    Percentage improvements in cost suboptimality (top) and number of SCP iterations (middle) relative to REL, along with comparisons of infeasibility rates (bottom), achieved by warm-starting the SCP using ART models trained on datasets with varying degrees of final time diversity. 
    Each bar represents the average improvement over all test samples corresponding to the final time~$T_f$ indicated on the x-axis.}
    \label{fig:final_time_diversity}
\end{figure}

To study the impact of diversity in the training dataset on the learning process, we repeat the aforementioned evaluation process for all the models presented in \cref{tab:training_diversity}. 
Concretely, these models differ in the number of scenarios used during training, ranging from a single scenario across all training trajectories (i.e., -1t and -1s), to a different scenario for every training trajectory (i.e., -Rt and -Rs).
Results for generalization to obstacle avoidance and time constraints are reported in \cref{fig:obstacles_diversity} and \cref{fig:final_time_diversity}, respectively\footnote{Results generated on a Linux system equipped with a 4.20GHz processor and 128GB RAM, and a NVIDIA RTX 4090 24GB GPU.}.
In \cref{fig:obstacles_diversity}, higher values on the x-axis represent higher complexity of the scenario---measured through the initial constraint-violation-budget associated to the REL solution, $\calC_{\mathrm{REL}} (t_1)$.
In \cref{fig:final_time_diversity}, results are aggregated according to the desired final time.
As expected, ART models trained on a single scenario \textit{or} final time are unable to effectively generalize across different scenarios and time constraints. 
Despite positive cost improvements observed for problem specifications close to the ones encountered during training (top section of \cref{fig:obstacles_diversity} and \ref{fig:final_time_diversity}), the performance of such models remains highly unreliable. 
This results in a strong degradation in terms of convergence speed and infeasibility rate (middle and bottom sections of \cref{fig:obstacles_diversity} and \ref{fig:final_time_diversity}). 
For instance, ART-1t exhibits an infeasibility rate of $\sim100\%$ when $T_f=40 \ s$.
Thus, it is expected that increasing the diversity of the training dataset would improve performance.
This hypothesis is validated in the context of obstacle avoidance generalization, where ART-Rs consistently outperforms the baseline and all other models, showing cost reductions between $10\%$ and $30\%$, convergence speed improvements between $7\%$ and $12\%$, and an infeasibility rate limited between $2\%$ and $8\%$. 
However, in the context of time constraint generalization, models trained on a diverse but discrete set of final times (e.g., ART-4t) maintain cost improvements and infeasibility rates comparable to ART-Rt while achieving faster convergence, with a steady $15 \%$ improvement in the number of SCP iterations across all final times). 
These results suggest that, while dataset diversity generally improves generalization performance, excessive diversity may have negative effects depending on the specific application. This is especially true for scenario variations that drastically alter the overall trajectory (e.g., longer final times allow the robot to follow the same trajectory in state space but with significantly reduced control effort).

\vspace{1mm}
\noindent \textbf{(3) How does the choice of scene representation influence generalization to novel scenarios?}

In our experiments, we employed a vector-space scene representation leveraging relative obstacle information. 
Specifically, at each timestep, we denote $\calX_{\vectorspace i}=\left[ \calX_{\vectorspace i}^{1},...,\calX_{\vectorspace i}^{(M)} \right]$, with $\calX_{\vectorspace i}^{(m)}\!=\!\left[ \br_i - \br_{O}^{(m)}, \lVert \br_i - \br_{O}^{(m)} \rVert_2 - R_{O}^{(m)} \right]$, and $\br_{O}^{(m)}$, $R_{O}^{(m)}$ being the coordinates of the center and the radius of the $m$-th obstacle expressed in $\globalRF$, respectively. 
This choice results in substantially improved generalization over using e.g., the absolute position of the obstacles.
Specifically, let us consider an alternative version of ART-1s, whereby we utilize only absolute obstacle information, i.e $\calX_{\vectorspace i}^{(m)}\!=\!\left[ \br_{O}^{(m)}, R_{O}^{(m)} \right]$. 
Statistical analyses indicate that the ART-1s model trained with absolute obstacle information substantially deteriorates its performance compared to its relative-observation counterpart---$+77\%$ average cost, $+49\%$ average number of SCP iterations, $+290\%$ average infeasibility rate.
These findings illustrate that relative scene representations are crucial to maximize the generalization capabilities of ART, even when trained on a single scenario.
More broadly, similar insights are likely to apply beyond obstacle representations. 
Therefore, selecting the most effective input representations should be a key consideration in the design of learning-based trajectory generation approaches.
 
\subsection{Experimental results}
\addtolength{\textheight}{-4.5cm}   
Finally, the effectiveness of the proposed framework is tested on the real free-flyer platform introduced in Sec. \ref{subsec:freeflyer_problem}.
Specifically, we employ the best models identified through simulations in Sec. \ref{subsec:simulation_experiments} to warm-start the SCP optimizer and we vary either the obstacles' configuration (ART-Rs) \textit{or} the final time constraint (ART-4t) of Problem (\ref{optimizationProblem2}). The planned trajectory is then tracked by a downstream PID controller with $\mathrm{K}_\mathrm{P} = \mathrm{diag}([2.0,2.0,0.2]), \mathrm{K}_\mathrm{D} = \mathrm{diag}([10.0,10.0,0.4]), \mathrm{K}_\mathrm{I} = \mathrm{diag}([0,0,0])$. 

\begin{figure}[tb]
\centering
\includegraphics[width=1\linewidth]{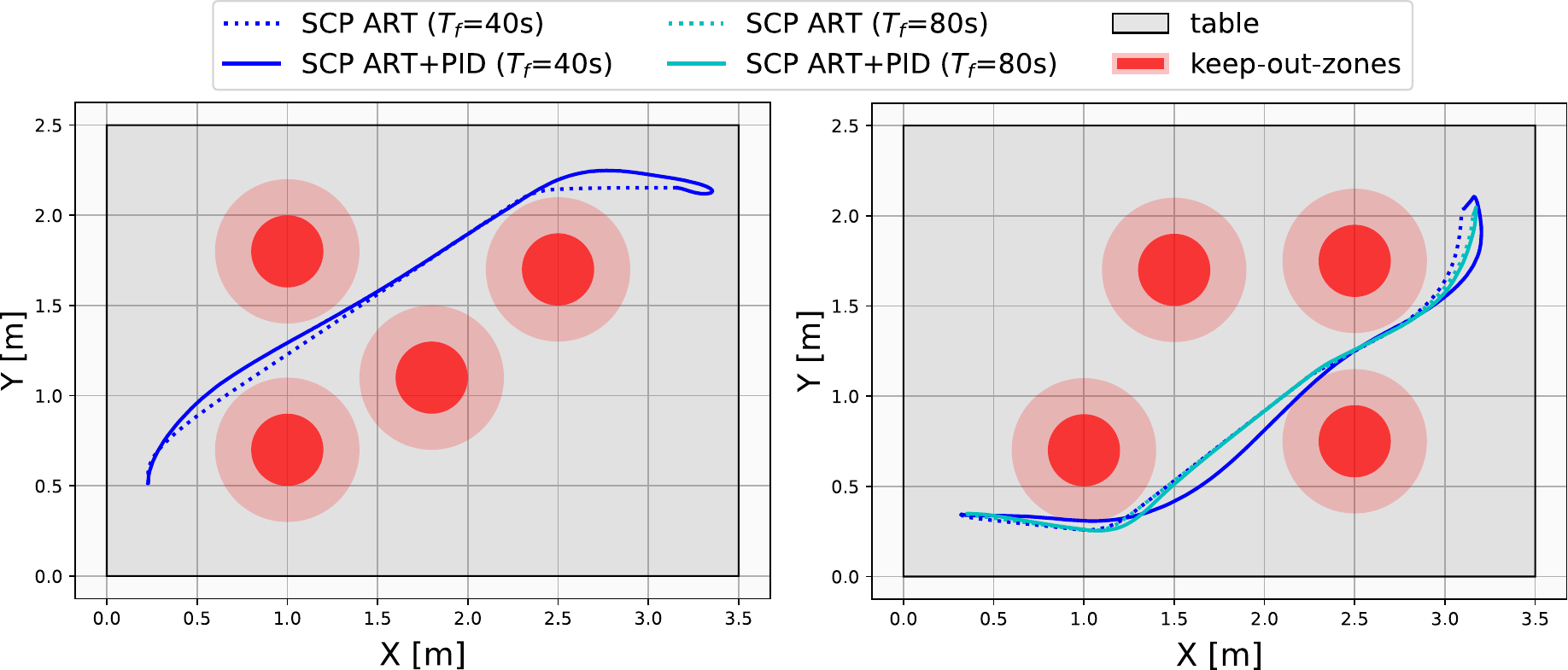}
    \caption{Trajectories planned using SCP ART (dashed lines) and tracked with a PID controller (solid lines) on the hardware testbed, varying obstacles' configuration (left) and final time (right). Videos: \href{https://acc25art.github.io}{https://acc25art.github.io}}
    \label{fig:hardware_trajectories}
\end{figure}
Qualitative results for both types of constraints generalization are shown in \cref{fig:hardware_trajectories}.
In the context of obstacle avoidance constraint generalization (left), we test ART-Rs in two of the three held-out scenarios introduced in Sec. \ref{subsec:simulation_experiments}. It is evident that ART-Rs demonstrates a robust capability to adapt the generation process in response to the current configuration of obstacles. This results in a final SCP solution that strategically leverages advantageous features of the scenario, e.g. the diagonal corridor shown in \cref{fig:hardware_trajectories} (left), rather than navigating through narrower spaces between obstacles. The cumulative firing time of all the eight thrusters of the platform is $ 88.34 s$. In the context of time constraints generalization (right), we assess the warm-starting performance of ART-4t utilizing two different final times, $40 \ s$ and $80 \ s$. 
In both cases, ART-4t provides an effective warm-start to the SCP, converging to two similar paths.
\begin{figure}[tb]
\centering
\includegraphics[width=1\linewidth]{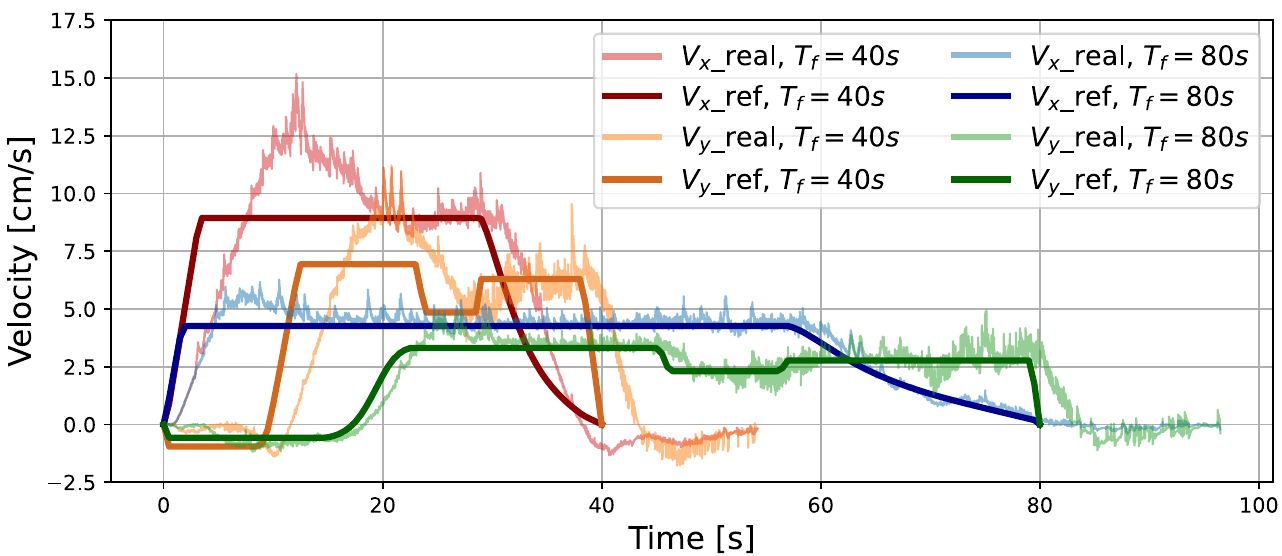}
    \caption{Reference and real velocity of the free-flyer along the $x$ and $y$ axes for the hardware experiments varying the final time constraint reported in \cref{fig:hardware_trajectories} (right).}
    \label{fig:hardware_velocity}
\end{figure}
The main differences between the two solutions is clearly visible in \cref{fig:hardware_velocity}, depicting the reference velocities computed by SCP ART for the free-flyer, as well as its real velocities achieved using the PID controller. Crucially, the specification of a larger final time constraint ($T_f= 80 \ s$) results in a less demanding trajectory characterized by lower translational velocities. This is further corroborated by the cumulative firing time of the thrusters of $90.54 \ s$ and $58.33 \ s$ for final time equal to $40 \ s$ and $80 \ s$, respectively.

\section{Conclusions}
In conclusion, this paper presents a novel trajectory generation framework which leverages the capabilities of transformer-based neural networks to address challenges related to multi-scenario spacecraft operations.
By integrating multimodal data into the learning routine, the proposed approach allows one to generate near-optimal trajectory guesses, improving the efficiency and robustness of traditional optimization methods.
Specifically, the experimental results showcased i) how the proposed framework can generate effective warm-starts for trajectory optimization in previously unseen scenarios, ii) the impact of data diversity, and iii) the impact of different scene representations on generalization capabilities.
The presented results highlight the potential of transformer-based models to overcome the limitations of \textit{single-scenario} trajectory generation frameworks, and underline their applicability to a broader range of space operations.
Future research interests primarily focus on three key areas. 
First, this research paves the way for extending the current framework to a broader class of complex trajectory optimization problems in spacecraft applications, such as in-orbit servicing, assembly, manufacturing, and logistics operations.
Second, enhancing the ART framework by introducing ad-hoc runtime monitors to provide stronger safety guarantees, particularly in detecting erroneous warm-starts from ART. 
Third, exploring the robustness of the proposed methodology in the presence of uncertainties, such as those arising from navigation, actuation, and unmodeled system dynamics.





\section*{Acknowledgments}
This work is supported by Blue Origin (SPO \#299266) as Associate Member and Co-Founder of the Stanford’s Center of AEroSpace Autonomy Research (CAESAR), the NASA University Leadership Initiative (grant \#80NSSC20M0163), the National Science Foundation (ECCS CPS project \#2125511), and the MIT Rudge (1948) and Nancy Allen Chair. This article solely reflects the opinions and conclusions of its authors and not any Blue Origin, NASA, or NSF entity.


\bibliographystyle{IEEEtran}
\small
\bibliography{ASL_Bib, main}

\end{document}

%% file: preamble.tex
\usepackage[acronym]{glossaries}
\usepackage[capitalise]{cleveref}

\theoremstyle{definition}
\newtheorem{example}{Example}

\newcommand{\opt}{\mathrm{Opt}}
\newcommand{\initcond}{\sigma_0}
\newcommand{\traj}{\tau}
\newcommand{\olloss}{\mathcal{L}}

\newcommand{\start}{\mathrm{start}}
\newcommand{\goal}{\mathrm{goal}}

\newcommand{\tup}[1]{\left( #1\right)}
\newcommand{\mat}[1]{\mathbf{#1}}

\newcommand{\dynapprox}{\hat{f}}

\newcommand{\cost}{J}

\newcommand{\statedim}{n_x}
\newcommand{\controldim}{n_u}

\newcommand{\bx}{\mathbf{x}}
\newcommand{\bX}{\mathbf{X}}
\newcommand{\bu}{\mathbf{u}}
\newcommand{\bU}{\mathbf{U}}

\newcommand{\bV}{\mathbf{V}}
\newcommand{\boldf}{\mathbf{f}}

\newcommand{\br}{\mathbf{r}}
\newcommand{\bv}{\mathbf{v}}

\newcommand{\horizon}{N}

\newcommand{\params}{\theta}

\newcommand{\calJ}{\mathcal{J}}
\newcommand{\calG}{\mathcal{G}}
\newcommand{\calR}{\mathcal{R}}
\newcommand{\calC}{\mathcal{C}}
\newcommand{\calX}{\mathcal{X}}
\newcommand{\calU}{\mathcal{U}}

\newcommand{\calS}{\mathcal{S}}
\newcommand{\calP}{\mathcal{P}}

\newcommand{\calT}{\mathcal{T}}
\newcommand{\FT}{\mathrm{FT}}
\newcommand{\OA}{\mathrm{O}}

\newcommand{\vectorspace}{\mathrm{V}}
\newcommand{\image}{\mathrm{I}}
\newcommand{\sens}{\mathrm{S}}

\newcommand{\real}{\mathbb{R}}
\newcommand{\integer}{\mathbb{N}}
\newcommand{\globalRF}{\mathrm{O}_{\mathrm{xy}}}
\newcommand{\bodyRF}{\mathrm{O}_{\mathrm{x}_\mathrm{b} \mathrm{y}_\mathrm{b}}}

\newacronym{acr:ml}{ML}{Machine Learning}
\newacronym{acr:av}{AV}{Autonomous Vehicle}
\newacronym{acr:rpo}{RPO}{Rendezvous and Proximity Operations}
\newacronym{acr:ocp}{OCP}{Optimal Control Problem}
\newacronym{acr:rl}{RL}{Reinforcement Learning}
\newacronym{acr:sl}{SL}{Supervised Learning}
\newacronym{acr:AI}{AI}{Artifical Intelligence}
\newacronym{acr:art}{ART}{Autonomous Rendezvous Transformer}
\newacronym{acr:mse}{MSE}{Mean-Squared-Error}
\newacronym{acr:llm}{LLM}{Large Language Models}
\newacronym{acr:scp}{SCP}{
Sequential Convex Programming}

%% file: main.bbl
\newcommand{\noopsort}[1]{} \newcommand{\printfirst}[2]{#1} \newcommand{\singleletter}[1]{#1} \newcommand{\switchargs}[2]{#2#1}
\begin{thebibliography}{10}
\providecommand{\url}[1]{#1}
\csname url@samestyle\endcsname
\providecommand{\newblock}{\relax}
\providecommand{\bibinfo}[2]{#2}
\providecommand{\BIBentrySTDinterwordspacing}{\spaceskip=0pt\relax}
\providecommand{\BIBentryALTinterwordstretchfactor}{4}
\providecommand{\BIBentryALTinterwordspacing}{\spaceskip=\fontdimen2\font plus
\BIBentryALTinterwordstretchfactor\fontdimen3\font minus \fontdimen4\font\relax}
\providecommand{\BIBforeignlanguage}[2]{{%
\expandafter\ifx\csname l@#1\endcsname\relax
\typeout{** WARNING: IEEEtran.bst: No hyphenation pattern has been}%
\typeout{** loaded for the language `#1'. Using the pattern for}%
\typeout{** the default language instead.}%
\else
\language=\csname l@#1\endcsname
\fi
#2}}
\providecommand{\BIBdecl}{\relax}
\BIBdecl

\bibitem{NesnasEtAl2021}
I.~A. Nesnas, L.~M. Fesq, and R.~A. Volpe, ``Autonomy for space robots: Past, present, and future,'' \emph{Current Robotics Reports}, vol.~2, 2021.

\bibitem{MalyutaEtAl2021}
J.~Alonso-Mora, S.~Samaranayake, A.~Wallar, E.~Frazzoli, and D.~Rus, ``Advances in trajectory optimization for space vehicle control,'' \emph{{Annual Reviews in Control}}, vol.~52, 2021.

\bibitem{RuggieroEtAl2023}
D.~Ruggiero and E.~Capello, ``Model predictive control for spacecraft swarm proximity operations,'' in \emph{{SICE Annual Conference}}, 2023.

\bibitem{RuggieroEtAl2024}
D.~Ruggiero, I.~Basnayake, H.~Park, and E.~Capello, ``Attitude and position control for formation flying of space robots equipped with a robotic manipulator,'' \emph{{Acta Astronautica}}, vol. 222, 2024.

\bibitem{silver2016mastering}
D.~Silver, A.~Huang, C.~J. Maddison, A.~Guez, L.~Sifre, G.~Van Den~Driessche, J.~Schrittwieser, I.~Antonoglou, V.~Panneershelvam, M.~Lanctot \emph{et~al.}, ``Mastering the game of go with deep neural networks and tree search,'' \emph{{Nature}}, vol. 529, 2016.

\bibitem{ParkEtAl2024}
\BIBentryALTinterwordspacing
T.~H. Park and S.~D'Amico, ``Bridging domain gap for flight-ready spaceborne vision,'' 2024. [Online]. Available: \url{https://arxiv.org/abs/2409.11661}
\BIBentrySTDinterwordspacing

\bibitem{ieee_2019_computing}
C.~Adams, A.~Spain, J.~Parker, M.~Hevert, J.~Roach, and D.~Cotten, ``Towards an integrated gpu accelerated soc as a flight computer for small satellites,'' in \emph{{IEEE Aerospace Conference}}, 2019.

\bibitem{GuffantiGammelliEtAl2024}
T.~Guffanti, D.~Gammelli, S.~D'Amico, and M.~Pavone, ``Transformers for trajectory optimization with application to spacecraft rendezvous,'' in \emph{{IEEE Aerospace Conference}}, 2024.

\bibitem{CelestiniGammelliEtAl2024}
D.~Celestini, D.~Gammelli, T.~Guffanti, S.~D'Amico, E.~Capello, and M.~Pavone, ``Transformer-based model predictive control: Trajectory optimization via sequence modeling,'' \emph{{IEEE Robotics and Automation Letters}}, 2024.

\bibitem{art_ieeeaero25}
Y.~Takubo, T.~Guffanti, D.~Gammelli, M.~Pavone, and S.~D’Amico, ``{Towards Robust Spacecraft Trajectory Optimization via Transformers},'' in \emph{{IEEE Aerospace Conference}}, 2025.

\bibitem{BanerjeeEtAl2020}
S.~Banerjee, T.~Lew, R.~Bonalli, A.~Alfaadhel, I.~A. Alomar, H.~M. Shageer, and M.~Pavone, ``Learning-based warm-starting for fast sequential convex programming and trajectory optimization,'' in \emph{{IEEE Aerospace Conference}}, 2020.

\bibitem{IzzoEtAl2019}
D.~Izzo, M.~M{\"a}rtens, and B.~Pan, ``A survey on artificial intelligence trends in spacecraft guidance dynamics and control,'' \emph{{Astrodynamics}}, vol.~3, 2019.

\bibitem{HovellEtAl2021}
K.~Hovell and S.~Ulrich, ``Deep reinforcement learning for spacecraft proximity operations guidance,'' \emph{Journal of Spacecraft and Rockets}, vol.~58, no.~2, pp. 254--264, 2021.

\bibitem{FedericiEtAl2022}
L.~Federici, A.~Scorsoglio, A.~Zavoli, and R.~Furfaro, ``Meta-reinforcement learning for adaptive spacecraft guidance during finite-thrust rendezvous missions,'' \emph{Acta Astronautica}, vol. 201, pp. 129--141, 2022.

\bibitem{ChenEtAl2021}
L.~Chen, K.~Lu, A.~Rajeswaran, K.~Lee, A.~Grover, M.~Laskin, P.~Abbeel, A.~Srinivas, and I.~Mordatch, ``Decision transformer: Reinforcement learning via sequence modeling,'' in \emph{{Conf.\ on Neural Information Processing Systems}}, 2021.

\bibitem{JannerEtAl2021}
M.~Janner, Q.~Li, and S.~Levine, ``Offline reinforcement learning as one big sequence modeling problem,'' in \emph{{Conf.\ on Neural Information Processing Systems}}, 2021.

\bibitem{ChiEtAl2023}
C.~Chi, S.~Feng, Y.~Du, Z.~Xu, E.~Cousineau, B.~Burchfiel, and S.~Song, ``Diffusion policy: Visuomotor policy learning via action diffusion,'' in \emph{{Robotics: Science and Systems}}, 2023.

\bibitem{FedericiEtAl2024}
L.~Federici and R.~Furfaro, ``Meta-reinforcement learning with transformer networks for space guidance applications,'' in \emph{AIAA Scitech Forum}, 2024.

\bibitem{HeEtAl2016}
K.~He, X.~Zhang, S.~Ren, and J.~Sun, ``Deep residual learning for image recognition,'' in \emph{{IEEE Conf.\ on Computer Vision and Pattern Recognition}}, 2016.

\bibitem{AlcanKyrki2022}
G.~Alcan and V.~Kyrki, ``Differential dynamic programming with nonlinear safety constraints under system uncertainties,'' \emph{IEEE Robotics and Automation Letters}, vol.~7, no.~2, pp. 1760--1767, 2022.

\end{thebibliography}
